\documentclass[graybox]{svmult}

\usepackage{mathptmx}       
\usepackage{helvet}         
\usepackage{courier}        
\usepackage{type1cm}        
\usepackage{makeidx}         
\usepackage{graphicx}        
\usepackage{multicol}        
\usepackage[bottom]{footmisc}

\usepackage{psfrag} 
\usepackage{subfigure}
\usepackage{stfloats} 

\usepackage{float}
\usepackage{skmath}

\usepackage{tikz}

\graphicspath{{img/}}

\makeindex             

\newcommand\copyrighttext{%
  This article {\em Making effective use of healthcare data using data-to-text technology} appears in the book entitled {\em Advanced data science for improved healthcare} edited by 
  Sergio Consoli, Diego Reforgiato Recupero and Milan Petkovic, published by Springer Heidelberg.
  }
\newcommand\copyrightnotice{%
\begin{tikzpicture}[remember picture,overlay]
\node[anchor=south,yshift=10pt] at (current page.south) {\fbox{\parbox{\dimexpr\textwidth-\fboxsep-\fboxrule\relax}{\copyrighttext}}};
\end{tikzpicture}%
}

\begin{document}

\title*{Making effective use of healthcare data using data-to-text technology}
\author{Steffen Pauws, Albert Gatt, Emiel Krahmer, Ehud Reiter}
\institute{Steffen Pauws, Emiel Krahmer \at Tilburg University \email{S.C.Pauws/E.J.Krahmer@tilburguniversity.edu}
\and Albert Gatt \at University of Malta \email{Albert.Gatt@um.edu.mt}
\and Ehud Reiter \at University of Aberdeen \email{E.Reiter@abdn.ac.uk} 
 }
%
%
\maketitle

\abstract*{Healthcare organizations are in a continuous effort to improve health outcomes, reduce costs  and enhance patient experience of care. Data is essential to measure and help achieving these improvements in healthcare delivery. Consequently, a data influx from various clinical, financial and operational sources is now overtaking healthcare organizations and their patients. The effective use of this data, however, is a major challenge. 
Clearly, text is an important medium to make data accessible. Financial reports are produced to assess healthcare organizations on some key performance indicators to steer their healthcare delivery. Similarly, 
at a clinical level, data on patient status is conveyed by means of textual descriptions to facilitate patient review, shift handover and care transitions. Likewise, patients are informed about data on their health status and treatments via text, in the form of reports or via e-health platforms by their doctors. Unfortunately, such text is the outcome of a highly labour-intensive process if it is done by humans. It is also prone to incompleteness, subjectivity and hard to scale up to different domains, wider audiences and varying communication purposes. Data-to-text is a recent breakthrough technology in 
artificial intelligence which automatically generates natural language in the form of text or speech from data. 
This chapter provides a survey of data-to-text technology, with a focus on how it can be deployed in a healthcare setting. It will (1) give an up-to-date synthesis of data-to-text approaches, (2) give a categorized overview of use cases in healthcare, (3) seek to make a strong case for evaluating and implementing data-to-text in a healthcare setting, and (4) highlight recent research challenges. }

\abstract{Healthcare organizations are in a continuous effort to improve health outcomes, reduce costs  and enhance patient experience of care. Data is essential to measure and help achieving these improvements in healthcare delivery. Consequently, a data influx from various clinical, financial and operational sources is now overtaking healthcare organizations and their patients. The effective use of this data, however, is a major challenge. 
Clearly, text is an important medium to make data accessible. Financial reports are produced to assess healthcare organizations on some key performance indicators to steer their healthcare delivery. Similarly, 
at a clinical level, data on patient status is conveyed by means of textual descriptions to facilitate patient review, shift handover and care transitions. Likewise, patients are informed about data on their health status and treatments via text, in the form of reports or via e-health platforms by their doctors. Unfortunately, such text is the outcome of a highly labour-intensive process if it is done by healthcare professionals. It is also prone to incompleteness, subjectivity and hard to scale up to different domains, wider audiences and varying communication purposes. Data-to-text is a recent breakthrough technology in 
artificial intelligence which automatically generates natural language in the form of text or speech from data. 
This chapter provides a survey of data-to-text technology, with a focus on how it can be deployed in a healthcare setting. It will (1) give an up-to-date synthesis of data-to-text approaches, (2) give a categorized overview of use cases in healthcare, (3) seek to make a strong case for evaluating and implementing data-to-text in a healthcare setting, and (4) highlight recent research challenges. }

\copyrightnotice

\section{Introduction}
\label{sec:1}

Note-taking in medicine stems back from the time of Hippocrates in Classical Greece, when physicians wrote down case histories in the chronological order of observed events, signs and symptoms. A physician learned from
the ailment of a patient simply by listening and writing down the history of events and sensations mostly felt and experienced by the patient~\cite{Reiser2009}. 
In the early 1800s, physicians started to produce and permanently keep free-format patient case records for teaching purposes and personal remembrance. These records were complete narratives reflecting physician style and personality. Only in the 1900s did structured forms of documentation start to emerge to support patient examination, laboratory results, nurse notes and the like~\cite{Siegler2010}. 


Since the advent of digitization, clinical practice and workflows in healthcare delivery have a fully electronic and standardized flow of communication amongst healthcare professionals. This communication shares findings on patient status including examination, diagnosis, prognosis and treatment outcome, but also supports entering medication orders or other physician instructions, submitting billings and following up with health insurers for receiving reimbursement of services rendered. As the first medical specialty experiencing disruptive 
digital change, radiology has a fully digitized clinical workflow including a standardized setup of networked computers and storage devices that are put into use for reporting and 
communication, mainly aimed at increasing workflow efficiency and patient throughput. 

Text is the preferred modality to convey patient findings in clinical practice. It has been shown that clinical staff makes better clinical decisions when exposed to expert-authored textual 
summaries compared to time-trend physiological data only~\cite{Law2005, Vandermeulen2007}. The need for text comes with a downside for healthcare professionals: the 
text needs to be produced by them. Indirect patient care such as report writing and administration takes up a considerable amount of time. For instance, some recent 
observational studies revealed that medical specialists in the hospital spend about 40\% of their time on administrative tasks~\cite{Sinsky2016,Wenger2017}. This chapter claims that a significant portion of professional text writing in healthcare can be taken over by computers by leveraging {\bf data-to-text 
technologies}, potentially freeing clinical staff from many administrative duties and making them available for direct patient care. In addition, data-to-text allows for consistent, fast and timely text 
writing,  because it is not susceptible to time pressure and subjectivity, which can negatively impact the quality of human-authored reports.

Data-to-text is a particular instance of {\bf Natural Language Generation (NLG)}, which is commonly defined as `the subfield of artificial intelligence and computational linguistics that is concerned 
with the construction of computer systems than can produce understandable texts in English or other human languages from some underlying non-linguistic representation of information'~\cite{Reiter1997}.
Though there is little room for ambiguity about the type of output produced by a data-to-text system, since it is textual, the input can change significantly from one application to the other, varying from times series, numerical data or aggregated statistics to images or video. A crucial strength of data-to-text techniques is that 
they can be 
tailored to an intended reading audience and/or serve a particular communicative purpose. The same kind of information can be provided to a medical specialist, 
a nurse or a patient, in each instance changing the precise content, the presentation order of the information, the language used and the tone of the text. In a similar vein, the style and language 
in the text can be designed to, for example, inform, convince or coach a reader. 

Research in data-to-text draws on computational models of human language production, as well as on algorithms for search and planning in artificial intelligence; to some degree it also draws on studies of human cognition 
and psycholinguistics. In recent years, there has been an increasing emphasis in NLG research on deploying machine learning techniques on large datasets and text corpora, which are 
increasingly available through digital publishing and social media. These have resulted in many successful on-demand data-to-text applications in finance, meteorology, news, sports, education and 
healthcare. A recent and comprehensive survey of the current state of the art in Natural Language Generation, and data-to-text in particular, can be found elsewhere~\cite{Gatt2018}.

In the following sections, this chapter introduces data-to-text technologies, with an emphasis on both existing and potential use cases for data-to-text in healthcare. 
We offer a strong case for assessing, evaluating and implementing data-to-text in healthcare settings and highlight recent research activities which have arisen from synergies with adjacent data science fields.

\section{Data-to-text technologies}
\label{sec2}

\begin{figure}[H]
\sidecaption
\includegraphics[width=0.6\textwidth]{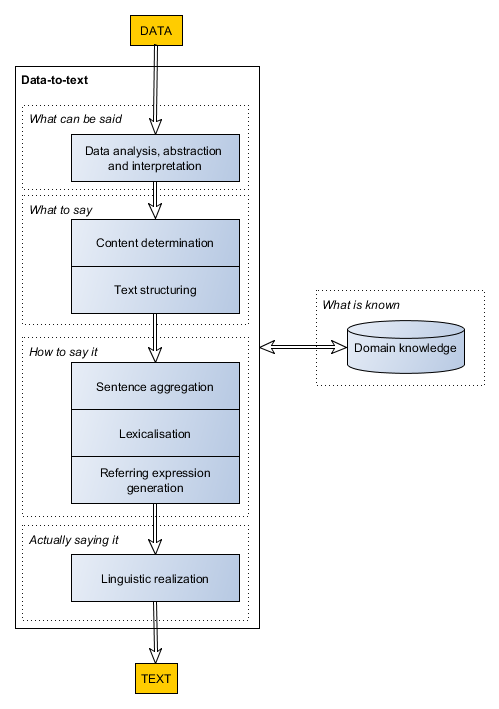}
%
%
\caption{Data-to-text pipeline.}
\label{pipeline-figure}       
\end{figure}

Traditionally, data-to-text systems make use of a pipeline of computing tasks 
to produce a coherent piece of text from input data~\cite{Reiter1997,Reiter2000}.
Roughly speaking and as shown in Figure~\ref{pipeline-figure}, these tasks focus on {\em what is known}, {\em what can be said}, on deciding {\em what to say} and {\em how to say it}, finally culminating in {\em actually saying it}.

\begin{itemize}
\item {\em What is known} involves the representation of domain knowledge for reasoning purposes. This serves as a common vocabulary in the application context that needs to be integrated in the 
whole functioning of the data-to-text system. Both domain knowledge and vocabulary can be kept in thesauri, taxonomies and ontologies, which are formal knowledge representations of medical 
concepts and their relationships. Some well-known ontologies in medicine are the Systematized Nomenclature of Medicine Clinical Terminology (SNOMED-CT\footnote{https://www.snomed.org/snomed-ct}) and those hosted by OBO Foundry\footnote{http://www.obofoundry.org/}.


\item {\em What can be said} involves the task of data analysis, abstraction and interpretation. This first core task is application-specific. Examples include the calculation of key performance indicators 
from medical billing data to be used in healthcare financial reporting, the interpretation of physiological sensor readings from bedside monitors in patient reporting, or the computation of risk and 
benefit estimates in patient data on diagnosis, treatment and outcome in shared decision-making.

\item {\em What to say} involves the task of content determination and text structuring. The former decides what information-bearing items will be presented in the output text based on the intended 
reader and communicative purpose. The latter decides the order of information-bearing items to be presented in the output text. 

\item {\em  How to say it} involves tasks such as sentence aggregation, lexicalization and referring expression generation. The first task decides how information-bearing items will be presented at individual sentence level, while the second one decides on what words and phrases will be used in expressing sentence-level information. The third task determines the content and form of phrases used to refer to domain entities, including pronouns and noun phrases.

\item {\em Actually saying it} refers to the actual text to be produced by means of linguistic realization. It produces a well-formed and coherent set of sentences as output text.

\end{itemize}

Below we use an example to elaborate on {\em What to say}, which is the task of content determination (Section \ref{subsec21}); {\em How to say it}, including both how to structure the text (Section \ref{subsec22}) and identifying the linguistic structures to express the content (Sections \ref{subsec23} -- \ref{subsec25}); and {\em Actually saying it}, which is usually referred to as linguistic realisation (Section \ref{subsec26}).

We will  not further discuss {\em What is known} and {\em what can be said} here, since these are similar to standard data science tasks.


\begin{figure}[H]
\sidecaption
\includegraphics[width=0.6\textwidth]{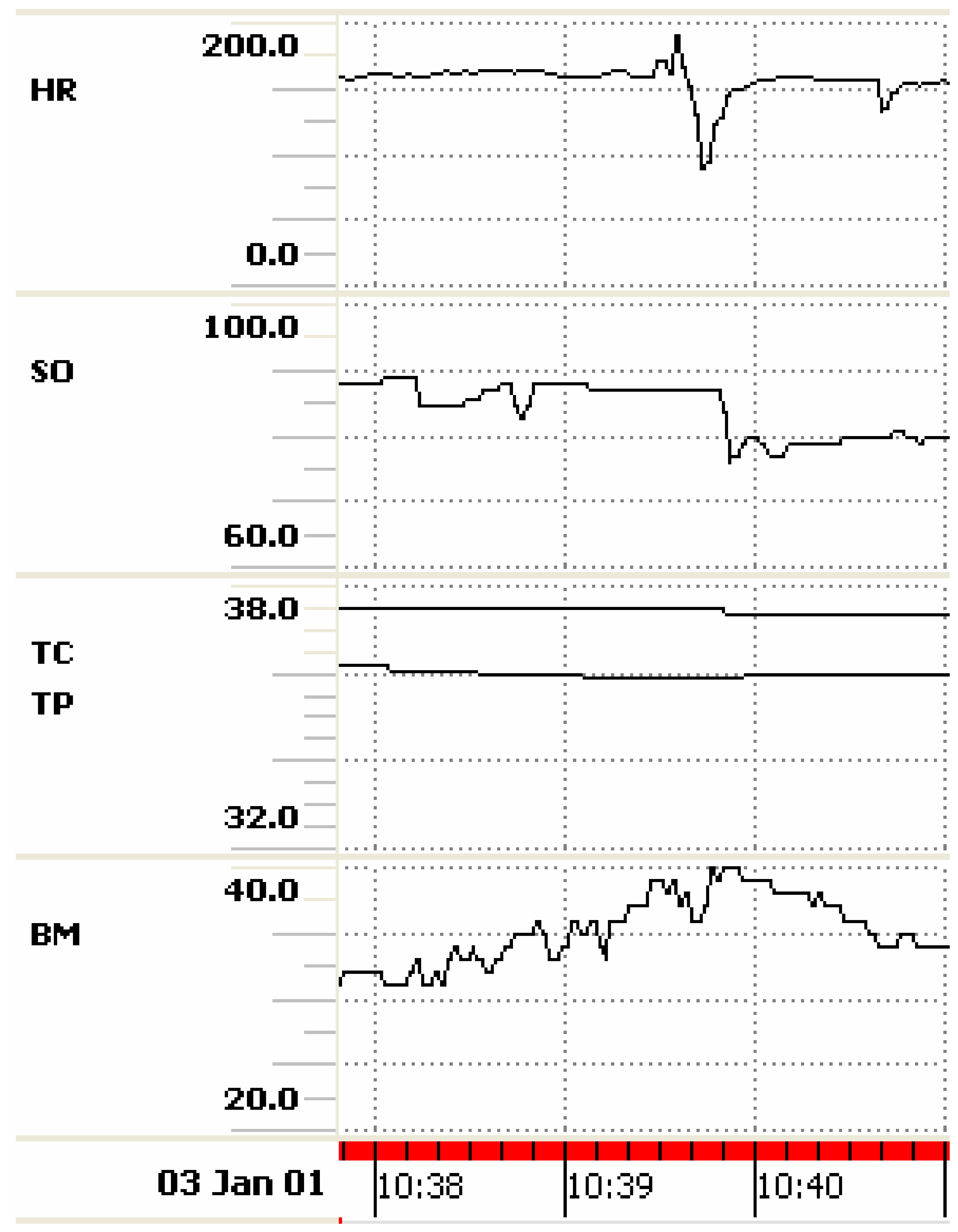}
%
%
\caption{Example time-series input data for Babytalk BT45 system (from~\cite{Reiter2007}). HR is heart rate, SO is oxygen saturation, TC is core temperature, TP is peripheral (toe) temperature and BM is mean blood pressure.}
\label{bt45-figure}       
\end{figure}

\subsection{Babytalk BT45 Example}
\label{btexample-sec}

We will use an example from the Babytalk BT-45 system \cite{Portet2009} (see Section~\ref{sec-intensive-care}), which generates summaries of 45 minutes of clinical data from babies in a neonatal intensive care unit (NICU), to support clinical decision making.  More specifically, we will look at the process of summarising the 3 minutes of example sensor input shown in Figure~\ref{bt45-figure} (of course a real summary would look at considerably more data), along with information about interventions in this period (in this example, morphine was administered to the baby at 10:39).

{\em What is known} in BT45 is based on a custom ontology of NICU events, interventions, etc.  BT45 determines {\em What can be said} (data analysis and intepretation) by applying signal processing techniques to segment the raw data and then using the ontology, together with rules collected from clinical experts, to label the events identified in the data with an index of their clinical importance. In the current example, the list of events identified is:

\begin{itemize}
\item Bradycardia (significant downward spike) in HR (heart rate), just before 10:40;
\item Desaturation (important downward step) in SO (blood oxygen saturation), again just before 10:40;
\item Upward spike in BM (mean blood pressure), at around 10:40;
\item TC (core temperature) is stable at 37.5$^{\circ}$C;
\item TP (peripheral temperature) is stable at 36$^{\circ}$C;
\item Morphine given to baby (intervention) at 10:39.
\end{itemize}

The output text produced by BT45 from this data is
\begin{quotation}
An injection of morphine was given at 10.39. There was a momentary bradycardia and mean BM rose to 40. SO fell to 79.
\end{quotation}

We discuss below the processing required to produce these 23 words from the input data.

\subsection{Content determination}
\label{subsec21}
Selecting what information to convey and what not to convey to the reader is key in achieving effective communication. In data-to-text, it is essential to strictly provide as much 
information as is needed, relevant and supported by the input data for the target reader, but no more. This, however, can depend on the communicative purpose. For example, it has been 
argued that a certain degree of redundancy can aid understanding, especially if the content being conveyed can be sensitive or distressing~\cite{Walker1992, DeRosis2000}. Information derived 
from input data is typically mapped to a pre-verbal representation, which can range from sets of attribute-value pairs, graph structures, schemas or any other convenient logical data structure. 

In our BT45 example, the content determination system decides that the text needs to mention the bradycardia, desaturation, upward spike in BM, and morphine administration, because clinicians making decisions should be aware of these events.  However it decides not to mention that TC and TP are stable, because these facts are less important in clinical decision making; hence the above BT45 text does not mention TC or TP.
This knowledge is encoded in the system in the form of expert system rules that allow the system to perform limited reasoning on the events identified in the data, assigning each event an index of importance. Importance may be pre-set (for example, certain events always have maximum importance and need to be mentioned) or it may be contextually determined (for example, a bradycardia may become more important in the context of a previous event involving the administration of morphine to the patient). Either way, importance values are used to decide what to say, resulting in a list of selected events (see above). These constitute the input to the text structuring stage.

\subsection{Text structuring}
\label{subsec22}

A text can be defined as a set of ordered and structured representations of information-bearing items. The optimal structure of a text often depends on the application domain. 
For example, many texts would start with a general introduction stating the general gist to be conveyed. Beyond that, the way information-bearing elements are grouped may be 
subject to genre constraints.  For example, if a text describes several events, these can be ordered by time, by what the events are about, or by how important the events are; the correct strategy depends on the domain and genre.  Sentence order also depends on relationships between sentences: we may want the text to mention the cause of some event before it mentions the event itself (the effect). 

A commonly used formalism for representing the relations, such as causality, contrast and elaboration, that can hold between items in a text, is 
Rhetorical Structure Theory (RST)~\cite{Mann1988}. The main idea behind RST is that 
items are represented as nodes in a labeled graph, whose edge labels indicate the relationships. In a data-to-text system, such a graph serves to structure information for the later modules responsible for fleshing out the text, in order to ensure that such relationships are adequately conveyed to the reader.

BT45 uses rules to structure the events selected by the previous stage using RST relations. These rules are domain-specific and applying them results in an RST graph whose nodes are the events themselves, linked via labelled edges, where the labels indicate relations ranging from causality (x caused y) to temporal sequence (x occurred prior to y), as well as groupings of events based on genre and domain considerations (x is linked to y because they are relevant to the same physiological system).

In our BT45 example, the text structuring system decides to mention the morphine event first, because it may have caused some of the other events (giving a baby medication can be stressful, hence leading to changes in heart rate, etc). This event is therefore linked to the others by a link specifying causality. The system also decides to mention the bradycardia and change in BM together, since these events are both about the cardiovascular system. 

The outcome of this stage is a labelled directed graph of events, which constitutes the input to the next stage.

\subsection{Sentence aggregation}
\label{subsec23}
Aggregation can be loosely defined as the process of reducing redundancy and enhancing the fluency of a text. Typically, this is done at sentence level, applying syntactic rules to 
merge sentential structures, thereby making the output more cohesive. For example, two sentences bearing the same subject (`The patient was intubated' and `The patient was 
given morphine') may undergo a rule 
to yield a single sentence (`The patient was intubated and given morphine'). While there is a substantial body of work on 
domain-specific aggregation approaches~\cite{Hovy1988,Shaw1998}, more recent work has sought to formulate generic algorithms that apply over syntactic structures 
independently of domain~\cite{Harbusch2009}. This implies that aggregation would be expected to apply later in the data-to-text process, after sentence planning has mapped content to 
syntactic structures. Other approaches formulate aggregation in tandem with content determination, jointly optimizing the choice and aggregation steps~\cite{Barzilay2006}, 
thus viewing aggregation as a pre-linguistic step.

Given the RST graph produced in the previous step, BT45 traverses the graph and determines which of the related events can be expressed in the same sentence. This too is performed using rules, which fire in response to specific types of event pairs in particular relations. In this case, 
the aggregation system decides to express the bradycardia and BM events in the same sentence ({\em There was a momentary bradycardia and mean BM rose to 40.}). This is because they are semantically linked (both about cardiovascular system), and expressing them in the same sentence highlights this linkage to the reader. Note that this is still a pre-linguistic decision. 

The outcome of this stage is a modified list of events, in which some may have been explicitly grouped as a result of an aggregation rule.

\subsection{Lexicalisation}
\label{subsec24}
Given a piece of information to be conveyed, an important step is to determine the words to use to convey it. 
Lexicalisation takes as input the list of events (some of which may have been aggregated) and the RST links between them, and brings their representations one step closer to their eventual linguistic expression.

Lexical choice can be driven by a number of considerations, including a 
target reader profile (which could, for example, determine the extent to which specialized terminology would be used~\cite{Mahamood2011}). Among the challenges in this task are the choice 
between near-synonymous terms which could nevertheless indicate subtle differences in meaning~\cite{Edmonds2002, Stede2000}, and the handling of vague expressions~\cite{VanDeemter2012}. 
Previous studies have suggested that in technical domains, such as weather forecasting, consistency in the use of lexical items is a valuable asset. For example, Reiter et al~\cite{Reiter2005} compared automatically generated weather forecasts for engineers to forecasts authored by meteorologists, and found that the former were preferred in part because they were more consistent in their use of vague temporal expressions. While meteorologists might exhibit slight variations in the meaning intended by an expression like `early evening', the system did not. This proved to be useful to the target users of the system.

While lexicalization is often discussed as an isolated task, many of the decisions taken at the word level will impact syntactic realization. For example, the same piece of information could be conveyed using 
the verb {\em give} or the verb {\em receive} (depending on the perspective from which an event is being viewed); this will have obvious repercussions on the realization of the sentence (compare: `doctors gave the patient morphine' vs. 'the patient received a dose of morphine'). For this reason, lexicalization is 
often viewed as part and parcel of the syntactic realization process ~\cite{Elhadad1997}.

In BT45, lexicalisation relies on a large lexicon of English, where words are also accompanied by information about their syntactic behaviour (for example, both `give' and `receive' require two arguments, a recipient and an agent, but their order is different and they need different prepositions to express the information). The lexicaliser uses rules that map concepts in the ontology to possible lexicalisations. At its simplest, this process simply chooses a noun or verb. However, additonal information in an event may also need to be expressed in words. For example, bradycardias are typically reported with an indication of their duration. Here, other rules need to be in place to map such information to the appropriate wording. In our running example,
it decides to express the length of the bradycardia using the vague word {\em momentary} instead of a numerical descriptor such as {\em 20-second} or an alternate vague descriptor such as {\em brief}. This was motivated by an analysis of how doctors described bradycardias in texts they wrote ~\cite{Reiter2008}.


At the conclusion of this stage, the original list of events has been fleshed out with lexical information, together with information about the syntactic frame in which the information needs to be expressed (e.g. {\sc recipient} {\em is given} {\em morphine} vs. {\sc recipient} {\em receive morphine from} {\sc agent}).

\subsection{Referring expression generation}
\label{subsec25}

Referring expression generation (REG) is a heavily studied field in computational linguistics~\cite{Krahmer2012}. When referring to an entity in text, it entails choosing a referential 
form such as a pronoun (`he'), a proper name (`Donald Trump') or a definite description (`the 45th President of the United States'). The choice of form is heavily dependent on the salience of a domain 
entity in the context, a fundamental observation in many discourse representation frameworks~\cite{Grosz1995,Poesio2004}. For example, a system may choose to 
refer to the president as `he' if he has recently been mentioned and there is no other, equally salient entity of the same gender with which the pronoun's intended referent might be confused. In case 
the intended referent is confusable with some distractors, it might be necessary to generate a full noun phrase (`the 45th US President' or `the man in the corner', for instance). This is actually a content 
determination problem: assuming some representation of the relevant features or properties of the referent, algorithms have been proposed to select a distinguishing subset of these properties~\cite{Dale1995}. Beyond pronouns and definite descriptions, natural language provides many other referential forms, including proper names. The choice of when to use proper names has only recently begun to be investigated in NLG or data-to-text~\cite{Castro-Ferreira2017,Siddhartan2011,VanDeemter2016}.

The REG module in BT45 operates over the lexicalised structures output by the previous stage. It identifies the entities (e.g. the {\sc recipient} of a drug), and decides how to express them in the text. 
In our example, the referring expression system decides to refer to the BM data channel as {\em mean BP} instead of {\em BM} or {\em mean blood pressure}.   If this channel was being referred to many times, the system could initially refer to it as {\em mean BP (BM)} and subsequently refer to it only as {\em BM}. This is based on genre conventions. The reference to this data channel resembles the use of a proper name (since there is typically only one blood pressure channel and it is a named entity). In the case of BT45, the use of names in such instances was largely heuristic in nature and guided by examination of example texts. Pronouns were used based on an estimate of the salience of the entity in the text: if an entity was mentioned previously, a pronoun might be generated instead of a name or a description.

At this stage, the original list of events identified during content determination has been fleshed out by identifying links between events, aggregating them, and fleshing out their lexical and syntactic properties, as well as referring expressions.

\subsection{Linguistic realization}
\label{subsec26}
Assuming that information-bearing units in a text have been selected, structured, and mapped to lexical representations, the final step in the traditional data-to-text pipeline involves realization. 
Lexical representations are mapped to syntactically well-formed sentences in the target language, a process that also necessitates handling morphological operations such as word 
inflections and subject-verb agreement, and inserting function words such as auxiliary verbs and complementisers. This is perhaps the sub-task of data-to-text for which there has been the greatest 
degree of development of `domain-independent' and reusable modules.

Perhaps the simplest approach to realization is template-based. Syntactic templates often take the shape of well-formed sentences with slots in which specific values can be filled in~\cite{McRoy2003}, 
although such templates can have a recursive form, making them potentially very expressive~\cite{VanDeemter2005}. 

Grammar-based systems tend to be much more complex. Realisers such as FUF/SURGE~\cite{Elhadad1996}, KPML~\cite{Bateman1997} or OpenCCG~\cite{White2012} involve a theory-driven 
description of the morphosyntax of a language, either using hand-coded rules or rules that are partially derived from treebanks. 

Increasingly, morphosyntactic realization is handled in a data-driven manner. For example, a realizer such as OpenCCG, or the earlier NITROGEN system~\cite{Langkilde2002}, might use language 
models derived from large corpora to select from among many possible realization options for the same input. Many of these systems rely on a chart algorithm as a base generator~\cite{Kay1996}, which 
produces multiple realisations of (parts of) input specifications and ranks them~\cite{Rajkumar2014}. Other approaches use classifiers to perform selection among options~\cite{Bohnet2010,Filippova2009}. 
A more recent turn, so far restricted to the generation of relatively short texts, involves the use of Recurrent Neural Networks as decoders to generate 
sentences directly, conditioning the generation on some non-linguistic input in an encoder-decoder framework~\cite{Gatt2018,Goldberg2017}.

In our BT45 example, the Simplenlg realiser \cite{Gatt2009b} is used to generate the final 23-word text,
as provided in Section~\ref{btexample-sec}.
This system provides an API for realisation where the decisions that drive the realiser need to be implemented directly by the developer. In BT45 this was handled by writing rules which deterministically mapped input structures (lexicalised events) to sentence structures. For example bradycardias are typically mentioned using `existential' constructions (`there is/was a bradycardia...'); hence, a bradycardia event would typically be realised by firing a Simplenlg procedure to produce such a sentence. Similarly, some events needed to be expressed in the passive voice. For example, this was always the case for drug administration events (`the patient was given morphine' rather than `the doctor gave the patient morphine'). Following the recursive application of these rules to map every part of a lexicalised input to a syntactic structure, Simplenlg's built-in functionality to handle morphology and agreement, as well as decide on capitalisation and punctuation, was applied and the final string was rendered, as shown above.

\section{Data-to-text in healthcare}
In this section, we present existing and potential use cases of data-to-text in healthcare and some of its adjacent fields. Use cases of data-to-text in healthcare are numerous as effective 
text-based communication is fundamental to ensure proper patient status sharing among clinical staff members. We  discern five different application areas, 
loosely referred to as {\em report automation} (Section \ref{5.1}), {\em clinical decision support} (Section \ref{5.2}), {\em behaviour change} (Section \ref{5.3}), {\em patient engagement} (Section \ref{5.4}) and {\em patient assistance} (Section \ref{5.5}). 

\subsection{Data-to-text for report automation}\label{5.1}
Report automation entails the automatic generation of routine text drafts that summarize statistics and findings. These text drafts can be edited by the end user before release, if desired. 
Automation saves time and increases accuracy and consistency in routine report writing. We discuss existing and potential use cases for healthcare finance, clinical practice, 
radiology, incidences during service and medical equipment utilization.

\subsubsection{Routine reporting on healthcare finance}

Financial reporting by means of data-to-text is already a viable commercial service; it includes annual, financial statements, investment and audit reporting primarily for banking and energy industry\footnote{See white papers of Arria at www.arria.com for further information on financial data-to-text services for banking and energy industry}. 
We conjecture that data-to-text will also be a mainstay for financial reporting in value-based health care ~\cite{Hunter2017a, Hunter2017b}. Due to new governmental legislation and health insurance policies, hospitals and health systems around the world are increasingly held financially accountable for keeping a healthy population in their catchment area, providing high quality services in case of sickness and improving patient experience. 
 The commercial success of data-to-text in financial reporting is mainly because financial reporting is a highly standardized and periodical mandatory prerequisite for external accounting compliance purposes. 
It retrospectively conveys financial standing of a hospital or health system over a specific period of time on beneficiaries, cost, profits and utilization statistics, while making sure the reported numbers are 
compliant to prevailing accountancy rules.

\subsubsection{Routine reporting in clinical practice}

Routine reports such as referral letters or patient examination findings are common in clinical practice. Current methods to produce these routine texts, such as the use of canned text or dictation, are far from optimal~\cite{Huske-Kraus2003a}.
The Suregen system~\cite{Huske-Kraus2003b} used data-to-text to assist
physicians in a hospital to write cardiology routine case reports. By using a Graphical User Interface (GUI), a physician indicated sign, symptoms and findings related to a patient suffering from heart disease, from which a case report was drawn up in the German language. 
In a related vein, Narrative Engine~\cite{Harris2008} assisted a general practitioner in generating legal narrative records of their 
patient encounters. Complete and accurate narratives are an important part of the patient record and are often used as legal records, for example in the context of  malpractice lawsuits. 

\subsubsection{Routine reporting in radiology}

Radiology 
is dominated by advanced 
imaging technologies and has fully embraced digitization. On request of a treating clinician, a specially trained physician interprets the images taken of a patient and produces a report containing the findings and 
diagnosis. Voice recognition (VR) dictation and conventional transcription services are the {\em de facto} method of report creation. Though performance of voice recognition has improved remarkably in last decade, 
there are ongoing challenges connected to it, including
production time, error and cost~\cite{Pezzullo2008}. 
Clinicians engaged in patient treatment often encounter a lack of clarity in a radiology report when key pieces of information need to be gleaned to plan patient care, 
as the reports come with great variability in language use, length, style and version~\cite{Ganeshan2018,Monico2010}. Structured reporting by means of report templates has been proposed as a way of
improving the quality of the reports. Related to that, there has been ample research on the preference of healthcare professionals regarding radiology report structure ~\cite{Travis2014}. 

Though there 
have been many proposals related to natural language and artificial intelligence technologies from market leaders in radiology reporting solutions, data-to-text has not yet been
considered in these proposals but is a potential use case. The tough nut to crack is the automated interpretation of medical imaging data and laboratory measurement data in the correct clinical context~\cite{Litjens2017}. Once this has been sufficiently accomplished the automatic generation of a report comes in naturally, done with immediate reference to the medical images (multimodality reporting, see also Section~\ref{6.2}) 


\subsubsection{Routine reporting on incidents}

A personal emergency response service (PERS) enables subscribers 
especially elderly people at risk of adverse events, to summon help at any time in situations that potentially require emergency ambulance 
transport to a nearby hospital. This can occur for a variety of reasons, such as 
a sudden worsening of a long-term condition, a fall incident or a sudden pain on the chest with shortness of breath. PERS involves a wearable device
such as a neck cord or 
wristband-style personal help button that, upon a button press, provides immediate contact with an agent in a 24/7 call center. The agent then dispatches the help request to an informal responder (e.g., a neighbor or 
family member) or calls an ambulance based on the subscriber’s situation.
The agents reassure the 
subscriber that help is on its way.  Follow-up calls are performed to assess the outcome of each incoming help request. Call center agents record unstructured and short-hand text notes during conversations with subscribers. Using these case notes, all help requests are classified according to their type, situation and outcome. A call center may serve many subscribers across disparate
geographical areas, on behalf of various healthcare or home care organizations. These receive reports on the incidents and help requests of their patients. In this context, data-to-text technologies can provide a consistent and efficient method of producing these reports from call center data. 

In a similar vein, data-to-text can be used for the purposes of generating Case Safety Reports (CSRs), for example, after people had adverse reactions to drugs. By making this information available in structured and easily readible format potentially helps to avoid future adverse reactions.

\subsubsection{Routine reporting on utilization}

Medical imaging technologies are capital investments of hospitals, for which strategic decisions need to be made on deployment, replacement and long-term financing. 
As an alternative to capital investment, under newer business models 
the equipment can be provided to a hospital by a lease arrangement at a relatively low cost, but with additional charges for the utilization of the equipment, estimated based on patient-hours.
Nevertheless, periodical routine reports on the 
inventory of the imaging technologies available, including volume, modalities, condition, maintenance and utilization of the equipment, helps a hospital to plan for strategic equipment decisions. Data-to-text is the 
pre-eminent technology to generate these routine reports. Past uses of such technology in related (albeit non-medical) settings include the generation of reports from time-series data collected in large gas turbines, for the use of experts during maintenance~\cite{Yu2005}.

\subsection{Data-to-text for clinical decision support}\label{5.2}

Clinical decision support (CDS) aims at assisting medical staff in making the right clinical decisions at point of care. We will discuss two existing use case from hematology and intensive care for neonates.

\subsubsection{Hematology}

Early data-to-text techniques were used in clinical decision support systems such as TOPAZ~\cite{Kahn1989}. TOPAZ summarizes blood cell counts and drug dosages of lymphoma patients over a period of 
time. A complete blood cell count provides an overview of the number and types of blood cells in a blood sample along with hemoglobin and hematocrit tests. It indicates the health status of a lymphoma patient 
before, during and after treatment. First, TOPAZ compares patient values on blood cell counts over time with population-based normal ranges for identifying deviations. Secondly, it groups deviating events into 
time intervals and searches for explanations. Lastly, it converts these explanations into text to be read by clinicians.

\subsubsection{Intensive care}
\label{sec-intensive-care}

In intensive care facilities, nurses are requested to provide a nursing report on patient observations and interventions at the end of their shift to facilitate hand-over and inform the treating physician. These 
manually authored reports often lack structure and can be rather biased due to subjectivity in interpreting a medical incident or due to the prevailing workload. Automatic generation of reports can overcome these limitations, 
notably because fast and timely generation technology is not susceptible to time pressure, which is one obvious reason why the quality of human-authored reports can suffer. Several pioneering data-to-text 
systems were developed for and tested in neonatal intensive care units (NICUs) under the rubric of the BabyTalk project (see also Section~\ref{btexample-sec}). The system described above BT45, was a pilot system that produced a nurse report by summarizing 45 minutes of historical physiological sensor data of admitted newborns, together with observations and records of interventions by the medical staff~\cite{Portet2009}. Off-ward tests in the NICU revealed that though BT-45 summaries did not match human-authored summaries in the quality of decision support they provided, they did yield comparable results to data presented using a visualisation, which is the standard way of presenting information in this context. Considering that the human-authored reports for each 45-minute segment used in the off-ward study took hours to produce, these results provided encouraging indicators on the feasibility of data-to-text technology in the NICU context.
The successor to BT45, BT-Nurse, summarizes twelve hours of live patient data~\cite{Hunter2012}
and was tested on-ward. During a 2-month on-ward live evaluation, the majority of the BT-Nurse summaries were found to be understandable, accurate and helpful, providing evidence 
that nurse report generation by computers is feasible and useful in clinical practice. Another system, BT-Family, generated summaries from patient data for parents of an admitted newborn. In this case, the system took into account the affective connotations of the information being presented, so as to 
reassure and help guardians
understand how their child is doing~\cite{Mahamood2011}. In an off-ward evaluation, parents who had previously had a preterm baby admitted to the NICU appreciated the affective language in the summaries.

\subsection{Data-to-text for behavior change}\label{5.3}

A person's health status is unmistakably affected by the person’s biology and genetics and the quality of healthcare available to that person in case of sickness. However, health is predominantly determined by person’s behavior and life style. Unhealthy lifestyles such as smoking, alcohol consumption, lack of physical activity, 
and poor access to healthcare results in increased risk of
mortality and morbidity~\cite{Ford2012}. Interventions to change `unhealthy' behavior in lifestyle are needed to
enhance longevity,
but they seem to have only limited impact.  We will discuss an existing use case in health promotion and a potential use case in recreational sports.

\subsubsection{Health promotion}
Health promotion concerns public policy on helping people to change their behavior 
to adopt 
a healthy lifestyle to prevent sickness later on in life. Smoking cessation is one of the foremost public health concerns, 
with promotion policies 
that target
taxation of tobacco, smoking restrictions in public areas, mass advertising campaigns, and health warning on tobacco products, among other measures. STOP is a data-to-text system 
designed to generate tailored smoking-cessation letters from data about an individual, acquired through
a four-page smoking survey~\cite{Reiter2003}. STOP was tested through a collaboration with general practitioners. In a clinical trial, tailored letters proved to be equally effective in motivating cessation as non-tailored letters,
though they led to 
a change in intent to stop in heavy smokers~\cite{Lennox2001}. Despite the lack of evidence that tailored letters yielded superior outcomes, letters did overall 
result in greater cessation rates, 
compared to a control group of 
participants 
who did not receive any letters~\cite{Reiter2003}.


\subsubsection{Sports}
Recreational sports practice, such as running in the park, can reduce risks for cardiovascular diseases and have mood-improving 
benefits~\cite{Lee2014}. However, any individual sports practice can have 
downsides on motivation, sustainability and responsible practice.  Recreational runners tend to set personal goals and targets for themselves, with only little follow-up after these targets have been met. While exercising for health purposes only, the 
enjoyment experienced may not be sufficient to guarantee that practitioners sustain the practice.
Finally, new and inexperienced sports people
are prone to risky or unhealthy practices, for example, by neglecting 
warming up, cooling down, or carefully pacing the exercise~\cite{Schiphof2017}.
This can result in injuries and early drop-outs. Data-to-text can generate personalized and persuasive coaching instructions from individual sports performance data, to be verbalized during exercising. Coaching instructions themselves are considered part of domain knowledge.


In addition, based on the performance data, data-to-text can tailor training schemes and draw up key points of attention for a healthy and responsible sports practice, 
thereby also enhancing motivation and the potential to sustain the practice over long periods. 
As the use of wearable devices to monitor performance during sporting activities increases, the potential for generation of personalised reports and messages is becoming ever greater. One application that has been piloted in a personal sporting context is ScubaText, a system to generate reports for scuba divers, to complement existing visualisation techniques~\cite{Sripada2007}.

\subsection{Data-to-text for patient engagement}\label{5.4}


Patient engagement refers to empowering patients in making their own choices in health and healthcare. Below we discuss decision aids, psychosocial plans and sleep quality as potential use cases for empowering patients.

Engagement starts with informing patients adequately about their health status and treatment options for building up trust between patient and doctor. Increasingly, patients demand better access to and more say about their data and treatments. If patients are more engaged with their health and treatment, it is generally assumed that better outcomes, higher patient satisfaction and lower costs in healthcare can be achieved. 

Technological advances, including data-to-text, offer the promise of automatically making health information more accessible for a broader range of patients and their relatives, by presenting information in a more personalised manner and by automatically adjusting the readability-level of the text to the intended audience. Since medical terminology is notoriously difficult for patients~\cite{Elhadad2006, Zeng-Treitler2008}, tools to automatically simplify and/or explain this terminology can  play a major role in conveying health information to patients with various levels of understanding~\cite{Zeng-Treitler2007}. 

Rephrasing words and sentences, or even modifying the text structure, as far as possible retaining the intended meaning, can be done using techniques such as paraphrasing~\cite{Madnani2010, Wubben2014}, simplification~\cite{Dras1999, Zhu2010} or compression~\cite{Cohn2007}. Many of these techniques are considered text-to-text (see~\cite{Androutsopoulos2010}, for example, for a survey of these techniques), where the input is (possibly complex) text and the output is (simplified) text. However, such techniques can also be combined with the data-to-text techniques discussed in the current overview. In general, such a combination of techniques can  be helpful for automatically generating different style-variations of a particular text (for example, simple as opposed to more complex, but also formal and matter-of-factly as opposed to informal and empathic), a topic to which we will briefly return in Section~\ref{6.4}.



\subsubsection{Decision aids}
Patients with a life-threatening diagnosis often face a difficult decision to choose from a range of treatment options with various outcomes and side-effects.  Physicians are obliged to inform patients about the chances of a favourable effect (long-term survival) and the risks of adverse effects (e.g., death, side-effects) of treatment options. In the case of cancer, various initial and adjuvant treatments are possible, such as surgery, radiotherapy, chemotherapy and hormone therapy, that may have  similar survival outcomes. However, the right therapy co-depends on patient preferences, since side-effects can affect cosmetics, sexual functioning, neuropathy and overall quality of life after survival. Only when well-informed, can patients participate in a shared decision making process with their doctor to discuss treatment options and preferences~\cite{Salzburg2011, Stiggelbout2012}. Decision aids assist patients in taking a role in shared decision making by providing relevant treatment options, explaining risks, benefits and outcomes for each option, exploring patient’s values and goals in life to elicit relevant preference and reaching a joint decision. 

Even though many decision aids have been developed already, their usefulness arguably falls short due their generic nature, being population-based and focusing exclusively on long-term survival, but above all lacking personalized explanations of health risks and benefits. A key question in the development of decision aids is how to present patient-specific information on risk and uncertainties in treatments in such a way that they can be appreciated and understood by the patient. 
The use of accessible language~\cite{Holmes-Rovner2011} as well as the combination of textual and visual explanations (multimodality)~\cite{GarciaRetamero2010, Spiegelhalter2011} have 
been proposed for better risk communication. Data-to-text allows decision aids to be automatically made, specific to the patient case at hand by generating text (possibly in combination with suitable visuals) that is tailored to the individual patient. 

PIGLIT was an early forerunner of a personalized decision aid. It dynamically generated hypertext pages explaining treatments to patient with cancer using the patient's medical record as the basis for personalization. In a trial, patients preferred the personalized information over general information~\cite{Cawsey2000}. In the same period, similar systems were created for migraine patients and diabetes patients. 

\subsubsection{Psychosocial plan}

Living with a chronic disease such as heart or lung failure comes along with coping with psychosocial problems as well. Patients wake up every morning knowing that they are sick, will get sicker in due course and never get well again.  In addition, the disease limits them in every-day physical abilities, enlarges their dependency upon family members and prescribes a strict medicalized way of living. Psychosocial factors play a major role in engaging patients in their treatment with respect to motivation, therapy adherence and lifestyle regimen~\cite{VanGenugten2016}. Therefore, patients are assessed on these psychosocial factors before treatment commences to find ways how to best coach patients to add high quality life to years. Report writing is an essential step to inform patients and instruct professionals about the coaching strategy. The need for tailoring, efficiency and consistency, data-to-text can take over the tedious role of professionals to write patient assessment and coaching reports. 

\subsubsection{Sleep quality}

Sleep apnea is a serious condition that reduces or stops breathing for several tens of seconds, at least five times per hour over-night. In addition to great fatigue during the day, it can lead to high blood pressure and even a stroke or a heart attack during sleep. Sleep apnea requires adjustment of the lifestyle such as smoking cessation or weight loss. An effective approach is Positive Airway Pressure (PAP) therapy in which a patient sleeps with a mask or cap on the mouth, nose or face and with an air pump device on the bedside table. The pump device provides a small air pressure to keep the airways open so that the patient can continue to breathe freely during sleep. The PAP therapy is only effective if the device is actually used. Therapy adherence 
is also a prerequisite to receive reimbursement from health insurance. Unfortunately, many patients struggle with the therapy or even stop the therapy early due to inconveniences of the mask, the hose and the pumping device.  As the pumping device collects treatment and usage data, tailored reports and visuals were produced by human copy writers in a study to inform patients on sleep quality, device use and coaching instructions if they encounter difficulties when sleeping with the device~\cite{Tatousek2016}. Such tailored feedback is crucial; from a total of 15,000 patients, patients receiving tailored feedback on their sleep had a therapy adherence improvement of 22\% and slept nearly one and a half hours longer than patients without such feedback~\cite{Hardy2014}. However, the tedious process of human report writing can be easily taken over by data-to-text.

\subsection{Data-to-text for patient assistance}\label{5.5}

Patient assistance refers to providing language tools to persons with communication disabilities to meet their undeniable needs in communicating with their social environment and relatives.

\subsubsection{Communication assistive tools}


Persons with severe communication disabilities such as voiceless locked-in patients already make use of communication assistive tools that support the communication of practical day-to-day goals and basic human needs such as hunger, thirst, discomfort and safety~\cite{Wilkinson2007}. Some of these systems allow for simple question and answering pairs with single words or short sentences. But to increase self-esteem of these persons, social interaction needs to be widened up towards truly engaging inter-personal communication, starting with telling a personal story of what  happened lately or has been in the news. However, producing a single sentence is extremely time consuming and exhaustive for this patient group which results in these patients being seldom engaged in social interaction. Data-to-text can generate  personal narratives based on sensor data of the still present limb or eye movements or the latest record of a person’s activities. For children with complex communication needs, a first data-to-text system named `How was school today?' exists and is in an evaluation stage. It produces personal kid stories from sensor data, photos and video to support interactive narratives about personal every-day experiences~\cite{Tintarev2016}.

\section{Evaluation of data-to-text in healthcare}

Evaluation of data-to-text systems has become a central methodological concern in NLG research~\cite{Dale2007,Gatt2018}. 
As a first fundamental methodological distinction, evaluations can be considered {\em intrinsic} or {\em extrinsic}. An intrinsic evaluation of a text assesses the linguistic quality or the correctness of the text, decoupled from the end user purpose of the data-to-text system. An extrinsic evaluation, on the other hand, assesses to what extent a data-to-text system is well-equiped to support the intended end user purpose (e.g., does it help in making a decision?)

Intrinsic evaluation can be done by asking human judges to rate text qualities such as {\em fluency} (e.g., `Does the text read naturally?'), {\em comprehensibility} (e.g., `Does the text have clarity?') or {\em correctness}/{\em fidelity} (e.g., `Does the text convey what it should convey regarding the input data?') while reading the texts. An intrinsic evaluation can take various forms: judges can be asked to indicate their preference  among different alternatives (e.g., both system generated texts and human written ones), or by rating texts with or without a human-authored reference. Another method of intrinsic evaluation relies on comparing the automatically generated text with human-authored texts using objective word-based metrics to assess the level of `humanlikeness' of the text. Precision and recall-related metrics such
as BLEU  (bilingual evaluation understudy)~\cite{Lin2004}, METEOR (Metric for Evaluation of Translation with Explicit ORdering)~\cite{Lavie2007} and ROUGE (Recall-Oriented Understudy for Gisting Evaluation)~\cite{Lin2003} can be used for this purpose. These metrics originate from, for example, machine translation (MT), and their application to NLG is not uncontroversial. 

With respect to extrinsic evaluation in the healthcare domain, we are concerned with the efficacy of data-to-text in being supportive to a particular end user task in a clinical workflow. In an evaluative setting, participants (i.e., prospective end users) are asked to accomplish these tasks either in a controlled experimental setting or on-site. A possibility in a controlled experiment is to have variants of generated texts or human-authored reference texts randomly assigned to participants, before conducting the tasks. Efficacy then relies  on an objective measure of task performance or achievement to indicate which text leads to better performance. The extrinsic studies of the BabyTalk systems are pioneering examples of experimental and on-site evaluation of data-to-text in a clinical setting~\cite{Portet2009, Hunter2012}. In case data-to-text acts as an interventional device to better health outcome, a randomized clinical trial (RCT) is considered a `gold standard' to demonstrate its efficacy. STOP was evaluated in a RCT to assess its effect on smoking cessation~\cite{Lennox2001}, and is still one of the few data-to-text systems evaluated in this way.

The breadth of intrinsic and extrinsic evaluation methods is more extensive than can be discussed here (see elsewhere for a more complete overview~\cite{Gatt2018}). In fact, finding out what the best way is to evaluate a data-to-text system, and especially, finding out how different intrinsic and extrinsic measures relate to each other is an important research challenge. 

\section{Research challenges}

In this section, we touch on some research challenges (in addition to evaluation) for data-to-text health applications: machine and deep learning (Section \ref{6.1}), use of multimodality (Section \ref{6.2}), temporal aspects (Section \ref{6.3}) and  stylistic variation (Section \ref{6.4}). More insights on general future directions in data-to-text can be found elsewhere~\cite{Gatt2018}.

\subsection{Machine and deep learning}\label{6.1}

Traditionally, data-to-text systems often relied on hand-crafted rules~\cite{Gatt2018, Reiter2000}. With more available data and computing power, data-driven approaches to text generation have become popular using machine learning and deep learning~\cite{Dethlefs2014, Gatt2018, Goldberg2016, Goldberg2017}. There is an interesting trade-off between these two approaches: rule-based approaches can generate output of a very high quality (essentially indistinguishable from human authored texts), but they are difficult to create and maintain and do not scale up well. Data-driven approaches, on the other hand, are more efficient and scalable, but the output quality may be compromised, due to the reliance of statistical information~\cite{Chen2008, Goldberg2017, Kondadadi2013, Konstas2013}. As a result, there is at the moment no guarantee that texts generated by the latter approach are always grammatically correct, accurate and easy to read. As a result, an important research question is how data-to-text systems can combine the strengths of the approaches, and none of the weaknesses. One promising line of future research involves hybrid approaches, which use statistical, data-driven approaches in limited, well-defined subtasks of data-to-text. 

\subsection{Use of multimodality}\label{6.2}


Multimodality refers to the combination of text and visuals in a single document, such as a radiology report in which findings expressed in text are embellished by cross-referencing to the medical images concerned. The integration of visuals in text and document presentation are largely overlooked subject areas~\cite{Power2003}. 
While textual presentation of clinical data is known to improve decision making, it is also well established that combining this with appropriate visuals can be even more effective. For example, one study revealed 
that the accuracy of decision-making by physicians is affected by both the type of graphical charts used and the framing of the clinical data~\cite{Elting1999}. In that study, icon displays and tables led 
to superior clinical decision making in comparison to pie charts and bar charts. Negatively-framed data led to better decision making than positively-framed data. 
If text is directly linked to visuals, it can enhance trustworthiness of the generated text, since the reader is able to cross-reference what is said in the text with what is visually represented. Key research questions are how a system can decide automatically which information to convey in text and which in images, and how images, like text, can be automatically generated in ways that make them easy to understand for readers.

\subsection{Temporal aspects}\label{6.3}

Textual summaries can extend over various time periods. For instance, a nurse report can disclose a 12-hour nursing shift or a weekly patient review. {\em Temporal aggregation} ensures that information can be textually
presented at various levels of detail, given the time period over which reporting needs to take place. Besides the linguistic component, aggregation also involves abstraction over the input data, for example, by 
creating a summarized description of physiological sensor time series data. Examples of the latter are `heart rate decreasing ending into bradycardia' or `saturation within target range for last hour'. Clearly, the 
number of potential reports that can be generated for different levels of aggregation over data quickly increases. Various researchers have addressed the problem of aggregation in data-to-text~\cite{Barzilay2006, Theune2006}, but it remains understudied and general solutions are lacking~\cite{Gatt2018}.

\subsection{Stylistic variation}\label{6.4}

Besides generating text from input data with high fidelity, data-to-text has focused over
the past ten year on the `stylistic variation' of the produced texts~\cite{Gatt2018}.
Varying in style allows the system to tailor the text to the reading audience or the communicative intent, while being provided with the same input data. For instance, the BabyTalk project was able to generate clinical summaries on newborns admitted in the NICU in formal, professional language to medical staff, in its BT-45 and BT-Nurse system, but also to produce text in informal, affective language for parents, in its BT-Family system. 
The challenge in 'stylistic variation' is first of all being able to operationalise the term 'style' in the actual use of words and grammar for a particular style. Second, varying in style implies that the data-to-text pipeline should be able to adapt (or  learn to adapt) to produce the desired stylistic effect.  A key challenge in data-to-text is developing systems that can indeed adapt their output to the intended audience and communicative intent, at all stages of the generation process.

In general, it seems fair to say that data-to-text techniques are eminently suitable to automatically deal with variation, in ways which would not be feasilble for human authors. This also generalises to, for example, multilingual generation, where the same data is expressed in different languages. This is an emerging theme within NLG, witness for example the recent multilingual surface (linguistic) realization task \cite{mille2017shared}. 

\section{Conclusion}


Data is increasingly important for many areas of healthcare, ranging from clinical diagnosis and decision support to patient empowerment and behaviour change. Text (possibly in combination with visuals) is the most preferred way of making data accessible, but this currently has a stumbling downside: healthcare professionals need to write these texts, which keeps them away from providing direct patient care. In many cases, however, these texts are part of a routine administration and follow a clear, well-defined structure, which raises the question whether the writing of texts cannot be automated. In this chapter, we have argued that data-to-text algorithms can indeed be used for this.

Data-to-text systems are capable of automatically converting input data into coherent natural language texts, using insight from computational linguistics and artificial intelligence. They can do this quickly, in large volumes and tailored towards individual readers. The quality of generated texts is high, at least on a par with texts produced by healthcare professionals working under time pressure and increasing work load. 

In this chapter, we have introduced the core tasks addressed by data-to-text system, and seen how they can be combined to form end-to-end systems converting input data into fluent output text. Moreover, we have surveyed a wide range of applications of these techniques in the health and healthcare domain, including both existing and potential future use cases. 

In recent years, data-to-text technologies have matured considerably, and are now commercially viable for the first time, in a range of application domains, including finance, weather, and media. The immense increase in available data and computing power as well as recent insights in data science and artificial intelligence have opened up exciting new opportunities for data-to-text in many areas of healthcare.

%
%
%

\end{document}